\documentclass[final]{cvpr}

\usepackage{times}
\usepackage{epsfig}
\usepackage{graphicx}
\usepackage{amsmath}
\usepackage{amssymb}
\usepackage{dsfont}
\usepackage{cuted}
\usepackage{capt-of}

\usepackage[pagebackref=true,breaklinks=true,colorlinks,bookmarks=false]{hyperref}

\begin{document}

\title{Single-Shot Freestyle Dance Reenactment}

\author{Oran Gafni\\
Facebook AI Research \\
{\tt\small oran@fb.com}

\and Oron Ashual\\
Facebook AI Research \\
{\tt\small oron@fb.com}

\and Lior Wolf\\
Facebook AI Research\\ and Tel-Aviv University\\
{\tt\small wolf@fb.com}}

\maketitle
\newcommand{\og}[1]{\textcolor{blue}{#1}}
\newcommand{\ogr}[1]{\textcolor{red}{#1}}

\begin{strip}\centering
  \centering
  \begin{tabular}{@{}c@{~~~~~~~}c@{~~~~}c@{~}c} 
 \includegraphics[height=4.23504cm]{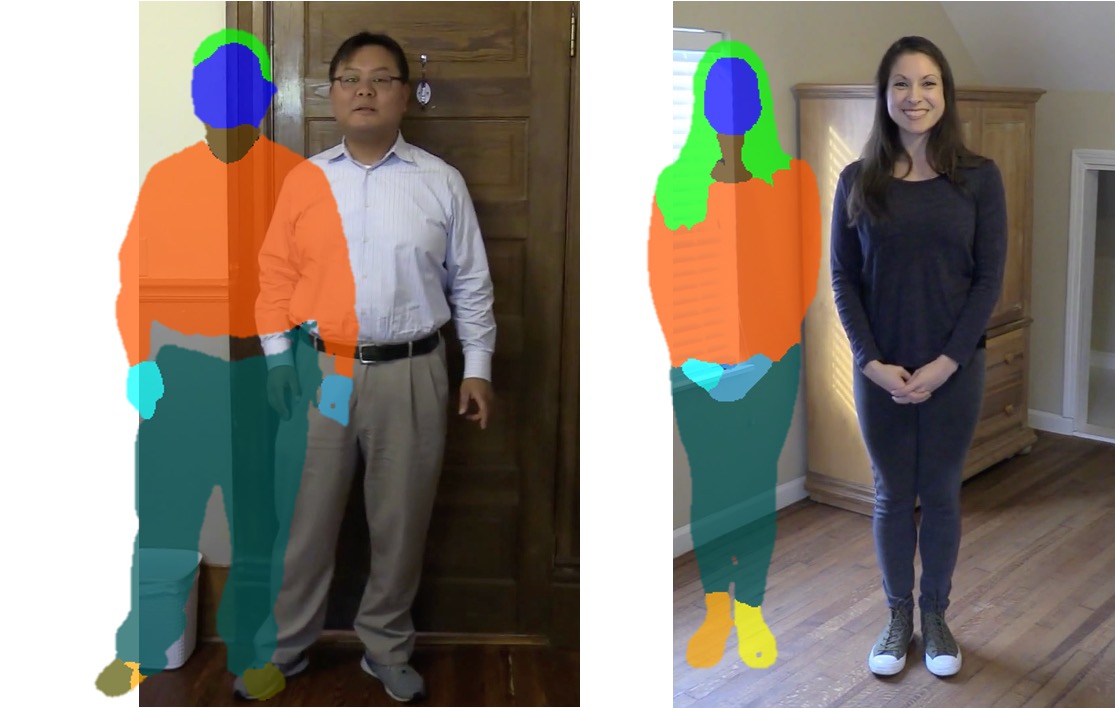} & 
 \includegraphics[height=4.23504cm]{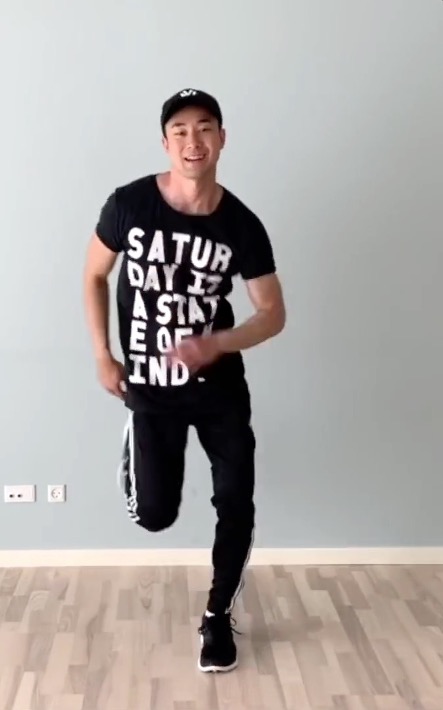} & 
 \includegraphics[height=4.23504cm]{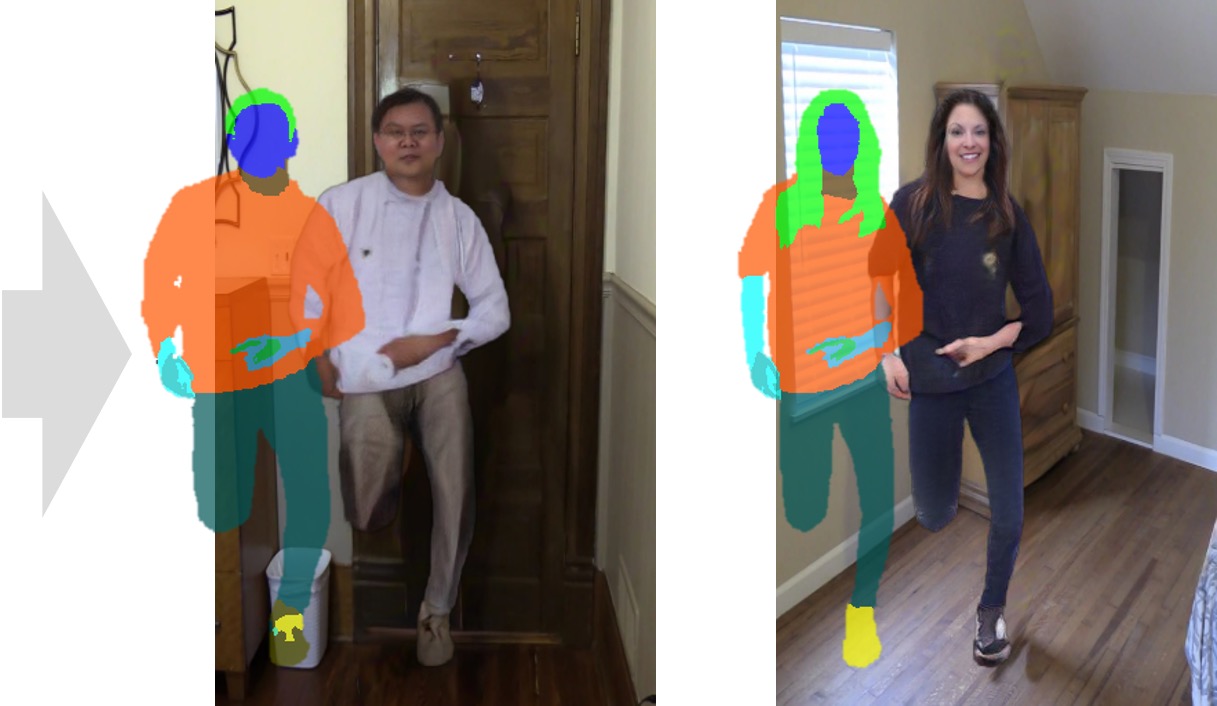} \\
 
 (a) & (b) & (c)
  \end{tabular}
  \captionof{figure}{Single-shot dance reenactment. {{ Using only a single image of a target person and their corresponding extracted semantic map (a), and a driving person's pose (b), we are able to render a novel corresponding semantic map of the target person, and a realistic person in the novel pose (c). Unlike previous work, we are able to maintain the body shape of the target person.}}}
 
  \label{fig:teaser}
\end{strip}

\begin{abstract}
The task of motion transfer between a source dancer and a target person is a special case of the pose transfer problem, in which the target person changes their pose in accordance with the motions of the dancer. 

In this work, we propose a novel method that can reanimate a single image by arbitrary video sequences,  unseen during training. The method combines three networks: (i) a  segmentation-mapping network, (ii) a realistic frame-rendering network, and (iii) a face refinement network. By separating this task into three stages, we are able to attain a novel sequence of realistic frames, capturing natural motion and appearance. Our method obtains significantly better visual quality than previous methods and is able to animate diverse body types and appearances, which are captured in challenging poses.
\end{abstract}

\section{Introduction}
The goal of this work is to animate a target person, who is specified by a single input image, to mimic the motion of a driving person, who is captured in a video sequence. This pair of inputs can be considered the easiest to obtain, and most minimalist and generic input for the given synthesis problem. Importantly: both the input image and the driving video are unseen during training.

The method we propose extends the envelope of the current possibilities in multiple ways: (i) the target person can vary in body shape, age, ethnicity, gender, pose, and viewpoint (ii) the sequence of poses that form the motion can be unconstrained, which is why we emphasize freestyle dance, (iii) the background can vary arbitrarily and is not limited to the source image or the background of the driving video.

This general setting contrasts with the limitations of existing methods, which often struggle to maintain the target person's appearance and avoid mixing elements from the driving video. The existing methods also often require an input video of the target person, have difficulty producing natural motion, and are limited to specific backgrounds. This is true, even for methods that train to map between specific persons seen during training.

To achieve this novel set of capabilities, we make extensive use of the latest achievements of neural networks for human capturing. Two pre-trained pose recognition networks are used to analyze the input video, a pre-trained human parsing network is used to segment the input image (of the target person), a pre-trained face embedding network is used to improve the face, and an inpainting network is utilized to extract the background of each training image. This maximal use of existing tools is an enabler for our method: using just one of the pose networks, or using pose in lieu of human parsing fails to deliver the desired results.

In addition to these components, for which there exist previous works that include a subset of it, we further employ specific representations. In order to ensure that the clothing {{and face appearance are}} captured realistically, we employ a five-part human encoder {{to the realistic frame-rendering network, consisting of four ImageNet-trained classifiers, and a trained face embedding network. These provide a rich embedding of the target, later enforced by a set of relevant perceptual losses}}. To ensure that finger motion is natural and the rendered hands do not suffer from missing parts, hand training data is augmented.

The method separates the pose and frame generation parts, performing each by a different network. The pose is provided in the space of a part-based segmentation map and is conditioned on both the target person and the motion frame. The second network transforms the generated pose and the target person's details to a masked frame, which is blended with an arbitrary background. The frame is further improved by applying a face refinement network based on an appearance preserving perceptual loss.

An extensive set of experiments is provided to establish the visual and numerical validity of the method. Compared to previous methods, our method provides considerably more accurate and visually pleasing results, as evaluated by a set of numerical metrics, a user study, and visual examples. Contrary to most previous work, we emphasize the ability to handle diversity in the target and generated individuals, promoting inclusion, which is generally lacking in this line of work.

\section{Related work}

A similar setting was presented in few-shot vid2vid (fsV2V)~\cite{wang2019fewshotvid2vid}, which generates a video sequence given a driver video and a source image containing a target person. Like our method, this method only trained once and can then be applied to any pairs of inputs. However, there are major differences in the applicability of the methods: our method can generate in arbitrary backgrounds, broader ranges of motions and is less restricting with respect to the inputs.  
Technically, fsV2V employs a hypernetwork~\cite{ha2016hypernetworks} that predicts the weights of a vid2vid network~\cite{wang2018vid2vid} given the target domain image(s), while our method employs conditioning based on this input. fsV2V suffers from flow-based artifacts, since it warps between consecutive frames, while our method generates entirely de-novo images. 
DwNet~\cite{siarohin2019first} also warps the input image based on the motion of the driver video.
Therefore, it is bound to the static background of the target person and suffers from artifacts around the animated character.

``Everybody dance now''~\cite{chan2018dance} and  vid2vid~\cite{wang2018vid2vid}, 
similarly to~\cite{wang2019fewshotvid2vid} generate an entire image, which includes both the target character and its background, resulting in artifacts near the edges of the generated pose~\cite{arbitrary,posetransfer}, background motion artifacts, and blurriness in some parts of the background. We employ a mask-based solution to integrate the generated character into an arbitrary background. 

Masks were previously used in the context of dancing to reanimate a specific person~\cite{zhou2019dance}. Methods that model a specific person do not need to model variation in body shape or capture novel appearances from a single frame.

Unlike our work and fsV2V, many methods require the target person to be specified by a video containing sufficiently varied motion (and not just an arbitrary still image), and are retrained per each pair of motion-source video and target-person video~\cite{chan2018dance,wang2018pix2pixHD,transmomo2020,ren2020human}. 

vid2game~\cite{gafni2019vid2game} is also trained per-person on a video containing a character's motion. 
Another difference from our work is that there is no replacement of appearance nor transfer of motion. Similar to our work, vid2game employes two networks Pose2Pose (P2P) and Pose2Frame (P2F), which are analog to two of the networks we use. However, the inputs and outputs differ from those of our networks, and the P2P network of vid2game generates similar poses in an autoregressive manner, while our task is more related to pose transfer. While vid2game is trained in a fully supervised manner, our network is trained in a self-supervised manner to reconstruct a person that exists in the image. 

Once the frame is obtained, we employ a face refinement network that utilizes an autoencoder architecture similar to the de-ID network~\cite{gafni2019live}. {{While~\cite{gafni2019live} seeks to distance the appearance from that of a target person, our method has opposite goals, bringing the appearance closer.}}

In still images, the problem of pose transfer is well studied~\cite{ma2017pose,siarohin2018deformable,balakrishnan2018synthesizing,wei2018person,dong2018soft,zhu2019progressive,esser2018variational,human3d,song2019unsupervised,men2020controllable, lwb2019}, out of which~\cite{dong2018soft,song2019unsupervised,men2020controllable} use a human parser, as we do. Most of these contributions employ images from the 
DeepFashion dataset~\cite{liuLQWTcvpr16DeepFashion}, which has four prominent disadvantages. First, the images posses a white background; second, the poses are limited to those encountered in fashion photography, and for example, the hands are rarely above the head; third, the body shapes are limited, and fourth, the number of different appearances, ethnicities and ages are few, resulting in overfitting to specific gender and age types. 

Another popular benchmark is the Market-1501 dataset~\cite{zheng2015scalable}, which depicts low-resolution images, with limited pose variability, that greatly differ from the dancing reenactment scenario. {{Explicit 3D modeling for single-image reanimation has been practiced as well~\cite{weng2019photo}, yet tends to result in unnatural motion and suffers from artifacts resulting from target image occlusions.}}

\section{Method}

Our method reenacts a character specified by a single input image, based on a given sequence of pose-frames. The method is designed to be generic, and the models are trained once and can then be applied, at test time, to any input character and motion sequence, without adjustments, re-training, or fine-tuning. 

The method relies on three image2image networks, each trained independently: (i) the P2B (Pose-to-Body) network maps pose and character information into body data, (ii) the B2F (Body-to-Frame) network maps the body-pose information obtained from the B2P and the character information to a frame, and (iii) the FR network refines the face in the frame generated by the P2F network.

On top of the three main networks we train (P2B, B2F, and FR), we employ an extensive set of pre-trained networks, in a manner that is unprecedented as far as we can ascertain: (i) a VGG network~\cite{simonyan2014very} trained on the ImageNet~\cite{deng2009imagenet} dataset that is used for obtaining the perceptual loss while training the B2F. (ii) A face detection and 2D alignment network~\cite{bulat2017far}. (iii) VGGFace2, which is a face embedding network~\cite{vggface2} that is used for training both the B2F and FR networks. (iv) The DensePose{~\cite{Guler2018DensePose}} network and (v) the OpenPose{~\cite{cao2018openpose}} network are both used to obtain pose information from each frame, as a way to represent the input of P2B. (vi) A human parsing network HP~\cite{li2019self} is used to extract the body in the target image. (vii) An inpainting network~\cite{yu2018generative,yu2018free} extracts the background from the training images, as well as  from the target image at inference time.

During training, we employ additional networks as discriminators that are denoted by $D_k$. There are a total of five discriminators: two are used for training the P2B, two for training the B2F, and one for training the FR.

The index $i=1,2,..$ is used to denote a frame index. The generated video frames (constructed from the output of B2F and FR) are denoted by $f_i$. The output of P2B is a sequence of generated semantic maps $P^M_i$ that are trained to mimic the output HP provides on real images of human figures. The input to P2B is comprised of two sequences: $P^D_i$ and $P^S_i$, denoting the dense annotation provided by DensePose, given a video $v$ and the stick figure and face landmarks output of OpenPose on $v$, respectively. In addition, P2B receives a semantic map $p^{M*}$ that denotes the parsing obtained by network HP for an input image $I$, that is used to specify the (target) person to reenact. 

B2F receives as input the sequence $P^M$ (here and below, the index is omitted to denote the entire sequence) and $e_z$, which is the concatenated embedding extracted by the pre-trained $\operatorname{VGGFace2}$ and $\operatorname{VGG}$ encapsulating the target person appearance. The output of P2F consists of two sequences:  $z_i$ denotes the generated image information, and $m_i$ is a sequence of blending masks (values between 0 and 1), that determines which image regions in the frame output would contain the information in $z_i$ and which would contain the background information provided by the user. The background information is denoted by $b_i$ and can be dynamic. The combination of the background with the synthesized images, in accordance with the masks is denoted by $f^0_i$. The output frames are generated by applying the refinement network FR to it.

Our method's flow consists of the following set of equations, given the input sequence of background frames $b$, image specification of the target person $I$, and a video containing the desired motion $v$.
\begin{align}
p^{M*} &= \operatorname{HP}(I)\\
P^D_i, P^S_i  &= \operatorname{DP}(v_i),  \operatorname{OP}(v_i)\label{eq:0} \\
P^M_i &= \operatorname{P2B}(p^{M*},P^S_i,P^D_i)\label{eq:1}\\
t_1, t_{2-5} &= \operatorname{l}(I,p^{M*}) \\
e_z &= [\operatorname{VGGFace2}(t_1),\operatorname{VGG}(t_{2-5})] \\
(z_i,m_i) &= \operatorname{B2F}(P^M_i,e_z))\label{eq:2}\\
f^0_i &= z_i\cdot m_i + b_i \cdot (1-m_i) \label{eq:3}\\
f_i &= \operatorname{FR}(f^0_i,t_1)\label{eq:4}
\end{align}
where $i=1,2,..$, $HP$, $DP$, and $OP$ are the Human-Parsing, DensePose and OpenPose networks respectively, the $P2B$ and $B2F$ are the Pose2Body and the Body2Frame networks. $l$ (Eq~\ref{eq:2}) is a function that separates the input image $I$ into 5 stacked 224x224 images $t_{1-5}$, containing the appearance of the (1) face and hair, (2) upper-body clothing, (3) lower-body clothing, (4) shoes and socks, and (5) skin tone, in accordance with the semantic parsing map $p^{M*}$. As stated, $B2F$ returns a pair of outputs, an image $z_i$ and a mask $m_i$ that are linearly blended with the desired background $b_i$ to create the initial frame $f^0_i$, using a per-element multiplication operator denoted by $(\cdot)$. FR takes this initial frame, and updates the face to better resemble the face of the target person, as captured in $I$. The semantic segmentation maps $P^M_i$ and $p^M_i$ are used in order to specify the face areas in the generated frame $f^0_i$ and in I, respectively.

\begin{figure}
  \centering
 \includegraphics[width=\linewidth]{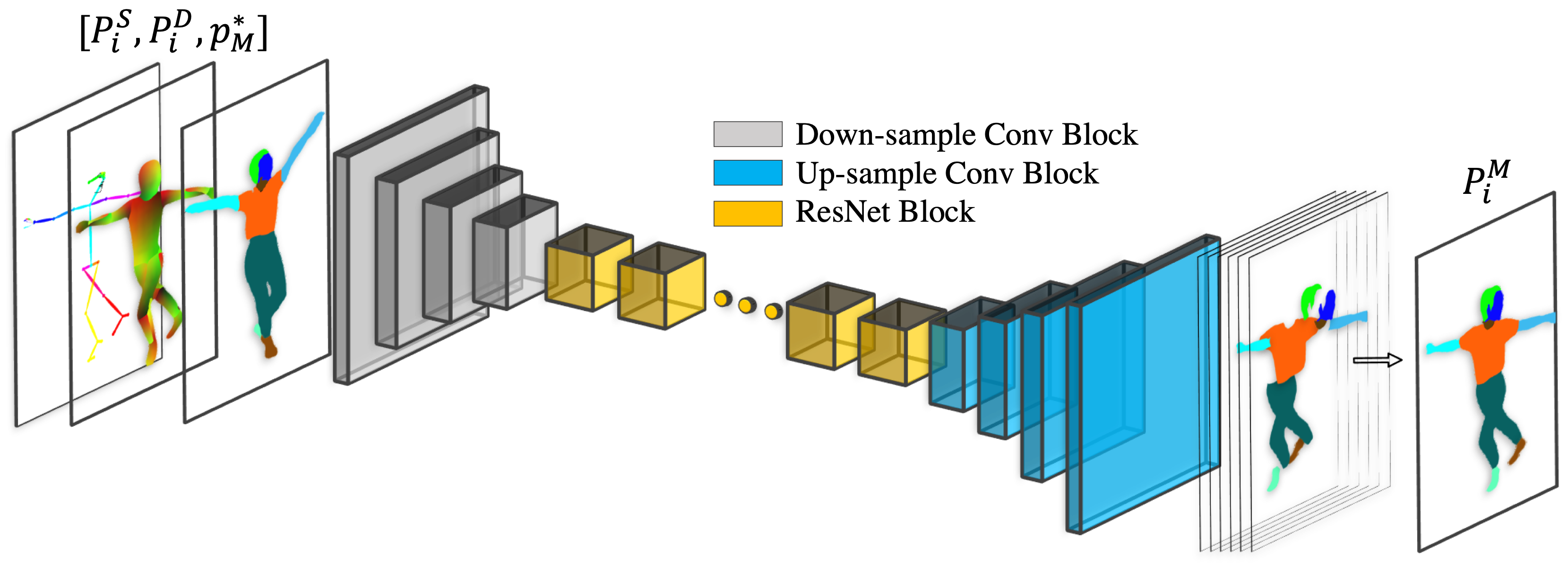}
 
  \caption{The architecture of the P2B network. Given a semantic segmentation of the target body $p^{M*}$, a source pose $P^S_i$, and a source dense pose $P^D_i$, the network generates the semantic map $P^M_i$ of the target person in the desired pose.}
  \label{fig:arch_p2p}
\end{figure}

\subsection{Pose2Body network}

The P2B's objective is to capture and transfer motion into the desired body structure, one frame at a time. The network has three inputs $p^{M*}$, $P_i^S$, and $P_i^D$. the first is produced by the human parser network applied to image $I$, the other two are obtained by pose networks, as applied to frame $i$ of the motion-driving video. The parsing map $p^{M*}$ consists of $22$ labels, of which $20$ labels are used as in the VIP dataset~\cite{zhou2018}, and $2$ labels are added to augment the hand landmarks extracted by OpenPose as labels. 

DensePose outputs three channels of the UV(I) space, where two channels project 3D mapping to 2D, and the third is a body index channel, with values between $0-24$. OpenPose generates key-points, which are joined to a single RGB stick-figure. Facial and hand landmarks are added to the stick-figure, increasing certainty and stability to the generated output. 

The P2B network utilizes the architecture of pix2pixHD~\cite{wang2018pix2pixHD}. In contrast to its original use for unconditioned image-to-image cross-domain mapping, we modify the architecture to allow it to generate a semantic segmentation map. Specifically, P2B produces the output $P^M_i$, which lies in the same domain as $p^{M*}$. 

The architecture of P2B is illustrated in Fig.~\ref{fig:arch_p2p}. Three inputs of the same spatial dimension are concatenated to one input tensor. The encoder part of the network is a CNN with ReLU~\cite{nair2010rectified} activations and batch normalization~\cite{ioffe2015batch}. The latent space embedding goes through $n_r$ residual blocks. Finally, the decoder $u$ employs {{fractional strided convolutions~\cite{figurnov2016perforatedcnns}}}, ReLU activations, and instance normalization~\cite{ulyanov2016instance}. A $\operatorname{sigmoid}$ non-linearity is applied after the last convolution to generate the output segmentation map.

\subsubsection{Training the Pose2Body network}

Following \cite{wang2018pix2pixHD}, we employ two discriminators (low-res and high-res), indexed by $k=1,2$. During training, the LSGAN~\cite{lsgan} loss is applied. An L1 feature-matching loss is applied over both discriminators' activations. In contrast to the B2F implementation, we apply a cross-entropy loss over the generated output.

The loss applied to the generator can be formulated as:
\begin{equation}
\begin{split}
\mathcal{L}_{P2B}=\sum_{k=1}^{2}{\left( \mathcal{L}_{LS^k} + \lambda_{D}\mathcal{L}_{FM_D^k}\right)} + \lambda_{CE}\mathcal{L}_{CE}
\end{split}
\end{equation}
where the networks are trained with $\lambda_{D}=40, \lambda_{CE}=1$.
The LSGAN generator loss is: 
\begin{equation}
\mathcal{L}_{LS^k}=\mathbb{E}_{(p^A_i)}\left[ \left(D_k(P2B(p^A_i))-\mathds{1}\right)^2 \right] \end{equation}
The expectation is computed per mini-batch, over the input HP, OP and DP $p^A_i=p^{M*},P^S_i,P^D_i$. The discriminator-feature matching-loss compares the ground-truth semantic map with the generated one, using the activations of the discriminator, and is calculated as:
\begin{equation}
\begin{split}
\mathcal{L}_{FM_D^k}=\mathbb{E}_{(p^A_i)}\sum_{j=1}^{M}\frac{1}{N_j}||D_k^{(j)}(P_i^M)- D_k^{(j)}(P2B(p^A_i))||_1
\end{split}
\end{equation}
with $M$ being the number of layers, $N_j$ the number of elements in each layer, and $D_k^{(j)}$ the activations of discriminator $k$ in layer $j$. The CE loss forces the generated 22 channels $P_i^{M}$ to be similar to the ground truth semantic map $P_i^{M*}$, and can be formulated as:  
\begin{equation}
\mathcal{L}_{CE}=CE(P_i^{M*},P2B(p^A_i))
\end{equation}

\begin{figure*}
  \centering
  \includegraphics[width=0.95\linewidth]{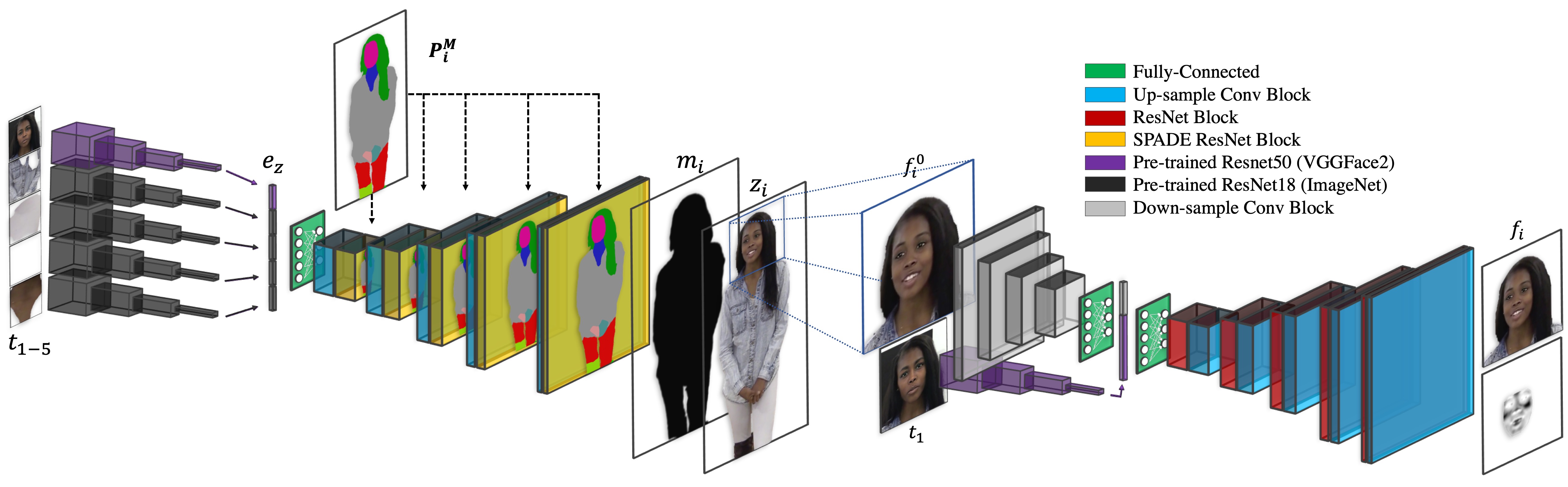}
  \caption{B2F and FR architectures. B2F receives as input the tensor $l(I,p^{M*})$ in which the segmented parts of the image $I$ {{are introduced through an array of pre-trained networks}}, and a conditioning semantic map $p^M_i$. The output frame $f^0_i$ is generated by blending a generated frame $z_i$ with the background $b_i$ in accordance with a generated mask $m_i$. FR extracts a face embedding utilizing a trained face embedding network and concatenated to the latent space. The pose, expression and lighting conditions are encoded for each input frame by the encoder, while the appearance can be taken from any image of that person.} 
  \label{fig:arch_p2f}
\end{figure*}

P2B is trained using the Video instance-level Parsing (VIP) dataset~\cite{zhou2018adaptive}. The dataset provides semantic segmentation annotations of people in diverse scenarios. Each training step relies on a single person in two different poses. To segment individuals in different views and poses, we rely on their location in a random frame, and an additional random frame, limited to a range of 250 consecutive frames. From the first, we utilize the semantic annotation, and DP/OP (Eq ~\eqref{eq:0}) as the network input, and the second is used for the semantic segmentation annotation ground truth, guiding towards the desired body-type and clothing.

\smallskip\noindent{\bf Disentangling body structure.} Few-shot generation methods suffer from the inability to generate a diverse set of body structures, as it is both challenging to correctly capture a body structure by a few samples, and datasets are highly biased towards certain body types. As a result, networks tend to learn a transformation of the source body structure, through the stick or dense pose representation, to the generated body structure. 

In addition to data augmentation in the form of random rotation and scaling of the inputs and output, we establish a more robust form of disentanglement between the guiding poses $P^S_i,P^D_i$ and the generated and source semantic maps $p^{M*},P_i^M$ , by introducing an additional form of data augmentation which is independent of the input and output body structures. We deliberately create a mismatch between the poses and semantic maps, by squeezing and stretching solely the body structures (segmentation maps) rather than the input poses. The network experiences samples that are in the exact same pose and view, yet differ in body structure. Examples of diverse body structure capability can be seen in {{ Fig.~\ref{fig:teaser} and in the supplementary.}}
\subsection{Body2Frame network}

B2F relies on two sources of input information: the generated pose of the target person $P_i^M$ and the encoding of the target person's image $I$. The latter is obtained based on image $I$ and its segmentation map $p^{M*}$. A stack $t_{1-5}=l(I,p^{M*})$ of five 224x224 images is created, corresponding to the resized bounding boxes around five semantic segments: (1) face and hair, (2) upper-body clothing, (3) lower-body clothing, (4) shoes and socks, and (5) skin tone. 

The output of B2F is a high-resolution ($512\times320$) frame $f^0_i$. The frames in the sequence $i=1,2,..$ are generated one by one, similarly to the P2B network. Each frame is generated by blending the background frame $b_i$ (can be static or dynamic) with the two outputs of B2F, the mask $m_i$ and the generated image $z_i$, as formulated in Eq.~\ref{eq:3}.

\smallskip\noindent{\bf Architecture.} The architecture of the B2F network is depicted in Fig.~\ref{fig:arch_p2f}. Image $t_1$ is passed through a pre-trained face embedding network to extract the appearance embedding, while images $t_{2-5}$ are encoded using a network pre-trained over the ImageNet dataset. The embedding extracted from the five pre-trained networks is concatenated into a single vector $e_z$ of size $2048+4*512=4096$. The latent space is projected by a fully connected layer to obtain a vector that is a reshaped tensor of size $4\times4\times1024$. The decoder has seven upsample convolutional layers with interleaving SPADE~\cite{park2019SPADE} blocks.

At test time, the latent space and FC layer are constant for a specific user, hence run only once, increasing the method's speed and applicability. 

\smallskip\noindent{\bf Datasets.} To enable diverse generation capabilities in terms of appearance (ethnicity, gender and age), pose, and perspective, we combine the Multi-Human Parsing (MHPv2)~\cite{zhao2018understanding,li2017towards} and the Crowd Instance-level Human Parsing (CIHP) {\cite{gong2018instance}} datasets. Both datasets contain various poses, viewpoints, and appearances, increasing the robustness of the network. Every annotated person is cropped to provide a single sample, that is later randomly resized for data augmentation purposes.

\smallskip\noindent{\bf Face emphasis.} Although a face refinement network is applied to the B2F output, it is limited in its refinement capabilities. Therefore, the B2F is required to generate a high-quality face as part of the novel person. The desired target face is introduced through the embedding, as extracted by the trained face embedding network. To encourage the generated face to be similar to the target face, both in quality and appearance, we apply a set of perceptual losses aimed at the expected position of the generated face. This is done in a pre-processing step, where all face locations are calculated using the face annotation. During training, these locations are adjusted to the random transformations applied, such as resizing, cropping, and flipping.

We apply a perceptual loss over the low, mid and high-level activations of a trained face embedding network. While high-level abstractions encourage appearance preservation, lower-levels handle other aspects, such as expressions. 
Additional guidance is provided to the face area in the form of explicit labels. Facial landmarks are used to draw five additional labels for the (1) eyebrows, (2) eyes, (3) nose, (4) lips, and (5) inner mouth. Although these landmarks are extracted from the driving (source) video, the perceptual losses applied to the face, as described in Eq.~\ref{eq:vggface}, help preserve the target person's appearance and expression.

\smallskip\noindent{\bf Blending mask.} B2F generates a blending mask in tandem with the generated character. This is imperative, as it enables the generated person to be embedded in any static or dynamic scene naturally. Training the B2F on an image dataset introduces an additional strain on the learning process of the blending mask, as there is no background image where the character is not present. To tackle this, we add a pre-processing step of inpainting all images, regenerating a region obtained by dilating the union of all semantic segmentation masks obtained by HP.
To increase generation quality, all losses are applied solely to the character. The semantic segmentation annotation labels are used to mask irrelevant image areas, such as the background or other people present in the same crop.

\paragraph{Loss terms} The following objective functions are used {{for training B2F}}:

\begin{equation}
\mathcal{L}^{G}_{hinge}=-\|D_{1,2}(P_i^M,z^b)\|_1 
\end{equation}
\begin{equation}
\begin{split}
\mathcal{L}^{D_{1,2}}_{hinge}=-\|\min(D_{1,2}(P_i^M,z^b)-1, 0)\|_1 - \\ \|\min(-D_{1,2}(P_i^M,x^b)-1, 0)\|_1  
\end{split}
\end{equation}
\begin{equation}
\begin{split}
\mathcal{L}^{D_{k=1,2}}_{FM}=\mathbb{E}_{(P_i^M,x^b,z^b)}\sum_{j=1}^{M}\frac{1}{N_j}||D_k^{(j)}(P_i^M,x^b) - \\
D_k^{(j)}(P_i^M,z^b))||_1
\end{split}
\end{equation}
with $M$ being the number of layers, $N_j$ the number of elements in each layer, $D_k^{(j)}$ the activations of discriminator $k$ in layer $j$, $z^b,x^b=z \odot P^{D+}_i,x \odot P^{D+}_i$, and ${L}^{G/D}_{hinge}$ as in~\cite{zhang2019self,lim2017geometric}.
\begin{multline}
\mathcal{L}^{VGG}_{FM}=\sum_{j=1}^{M}\frac{1}{N'_j}||VGG^{(j)}(x)-VGG^{(j)}(o))||_1
\end{multline}
with $N'_j$ being the number of elements in the $j$-th layer, and $VGG^{(j)}$ the VGG classifier activations at the $j$-th layer.

The network also outputs a mask, which is trained using the L1 loss to reconstruct a binary version of the HP frame $P^M_i$ after threshold at zero, denoted by $P^{D+}_i$ ($\lambda_m=5.0$):
\begin{equation}
\mathcal{L}^{m}_i=\lambda_m\|m_i - P^{D+}_i\|_1
\end{equation}

\subsection{Face refinement network}

The third network, FR, receives two inputs: the aligned face of the target person, as extracted from $I$, and the aligned face in the generated frame $f^0_i$. In both cases, the face is extracted and aligned using the method of~\cite{bulat2017far}.

The face crop obtained from $f^0_i$ is denoted $c^0_i$ and serves as the input to FR. The face crop obtained from $I$ and $p^{M*}$ is denoted by $c_I$, and it serves as a conditioning signal to this network. For this purpose, the pre-trained VGGFace2~\cite{vggface2} network is used, and the activations of the penultimate layer, denoted by $\operatorname{VGGFace}(c_I)$ are concatenated to the latent representation given by the encoder part of FR.

FR has the same autoencoder architecture as the de-id network~\cite{de-id}, which solves the de-identification problem, which is very different from the current face refinement goal. We, therefore, employ a perceptual loss that differs from that of~\cite{de-id} and minimize the following loss: 
\begin{equation}
    L_\text{facep} = \sum_j \|\operatorname{VGGFace}_j(c_I)-\operatorname{VGGFace}_j(c^0_i)\|
    \label{eq:vggface}
\end{equation}
where the index $j$ is used to denote the spatial activations size at specific layers of network VGGFace, and the summation runs over the last layers of each block of size $112\times112$, $56\times56$, $28\times28$, $7\times7$, $1\times1$ ($1\times 1$ being the size of the top-most block, i.e., $\operatorname{VGGFace}(c)=\operatorname{VGGFace}_{1\times1}(c)$). The rest of the loss terms (reconstruction losses, mask regularization losses, adversarial losses) are the same as~\cite{de-id}.

FR outputs a generated crop $c$ and a blending mask $m^c$:
\begin{equation}
    \begin{split}
        [c,m^c] = FR(c_I,c^0_i)\\
    \end{split}
\end{equation}
To create the final frame $f_i$, the crop $c$ is blended with the region of frame $f^0_i$ that corresponds to the face, in accordance with the values of the mask $m^c$.

\begin{figure*}
  \centering
 \includegraphics[width=0.93\textwidth]{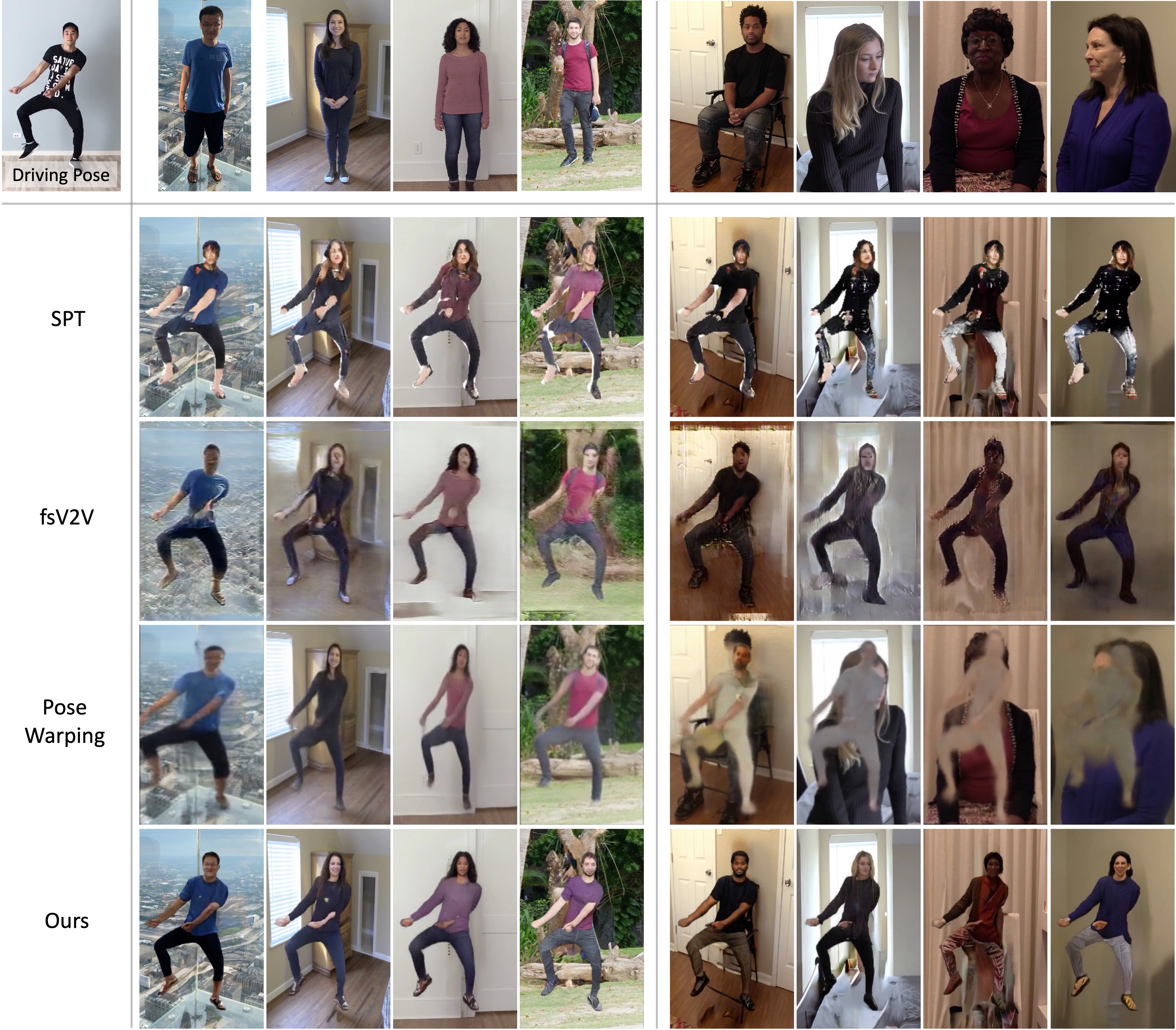}
 
 \smallskip
  \caption{Comparison with previous methods. {{ Each column presents a different target image. Our method is better able to handle both the "simple" (4 left) and "challenging" (4 right) targets, rendering higher quality and better appearance preserving results.}}}
  \label{fig:previous1}
\end{figure*}

\begin{table*}
\begin{center}
\begin{tabular}{|l|c|c|c|c|c|c|c|c|c|}
\hline
Method & SSBS $\uparrow$ & SSIS $\uparrow$& DPBS $\uparrow$& DPIS $\uparrow$& LPIPS 	$\downarrow$& LPIPS $\downarrow$&  SSIM $\downarrow$ & FID $\downarrow$ & Human \\
       &      &      &      &      & (VGG) &(SqzNet)&      &     & Preference \\
\hline\hline
fsV2V\cite{wang2019fewshotvid2vid}  & $0.870$ & $0.193 $ & $0.896$ & $0.436$& $0.567$ & $0.474$ & $0.255$ & $201.82$ & 0.98 \\
Pose Warping\cite{balakrishnan2018synthesizing} & $0.764$ & $0.143$  & $0.791$ & $0.347$ & $0.462$ & $0.372$ & $0.132$ & $159.71$ & 0.88 \\
SPT\cite{song2019unsupervised} & $0.851$ & $0.165$ & $0.862$ &$0.404$ & $0.378$ & $0.289$ & $0.127$ & $109.13$ & 0.81 \\
Ours & $\textbf{0.902}$ & \textbf{0.218} & $\textbf{0.928}$ & $\textbf{0.500}$ & $\textbf{0.375}$ & $\textbf{0.283}$ & $\textbf{0.116}$ & $\textbf{83.95}$ & - \\
\hline
\end{tabular}
\end{center}
\caption{Comparison with previous work. The last column denotes the percent of samples in which the users preferred our results over the baseline. {{All results were obtained on ''simple'' targets only, as previous methods could not handle ''challenging'' targets.}}}
\label{tab:results}
\end{table*}

\begin{figure*}[t]
  \centering
  \begin{tabular}{@{}c@{~}c@{~}c@{~}c@{~}c@{}}
  \includegraphics[height=3.52cm]{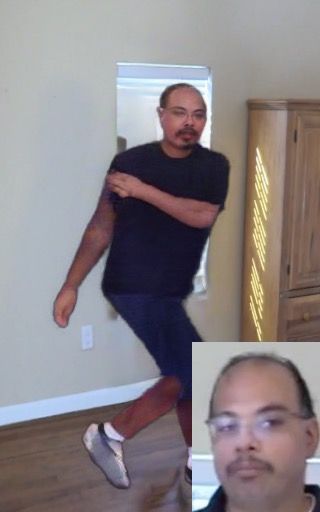} & \includegraphics[height=3.52cm]{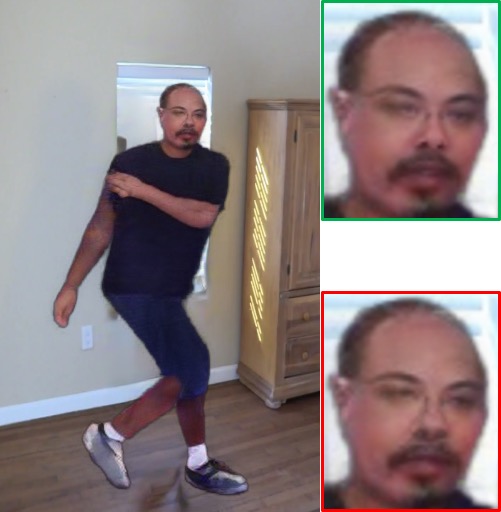} & 
  \includegraphics[height=3.52cm]{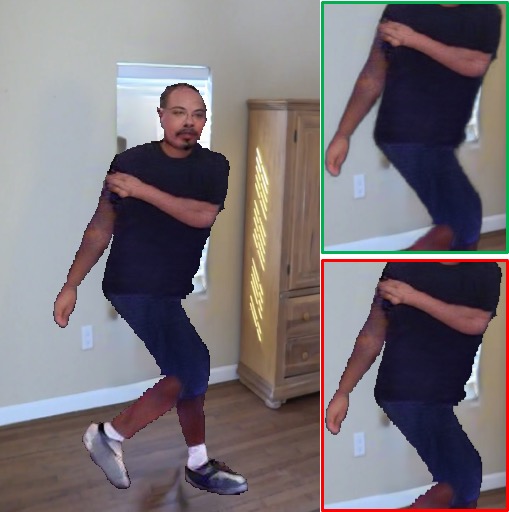} & \includegraphics[height=3.52cm]{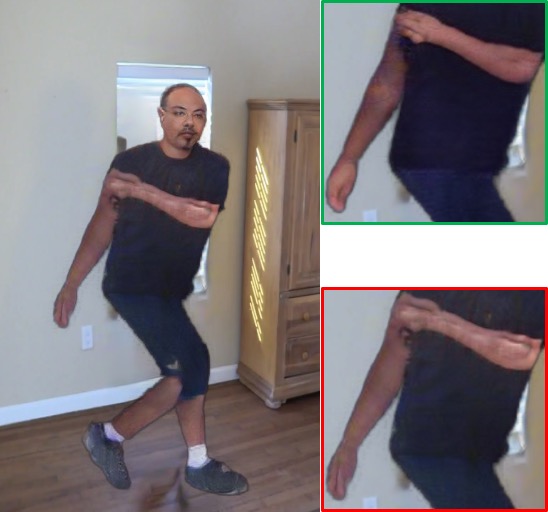} & \includegraphics[height=3.52cm]{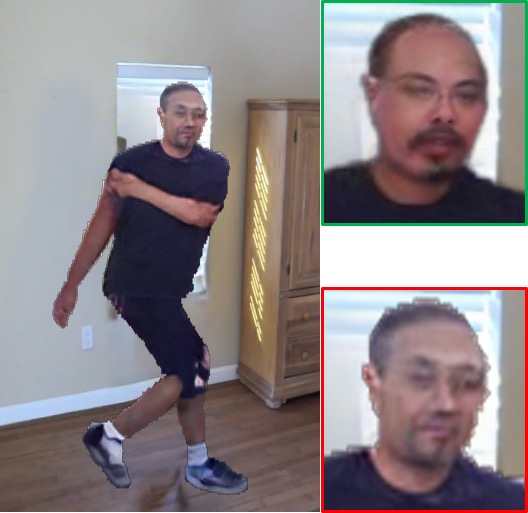} 
 \\
  (a) & (b) & (c) & (d) & (e)
  \end{tabular} 
  \caption{{{B2F/FR ablation study. (a) Our result and the target face. In the following, we show the resulting frame of a variant of our method. In red a zoom in of a certain part, and in green the same crop from our full method.  (b) No FR (\textit{blurrier face, features are less distinctive}), 
(c) no blending mask (\textit{crude edges surrounding the entire rendered character}),
  (d) hand/finger labels not added (\textit{arm distortions due to finger uncertainty, fingers less distinct}), (e) no face loss, lower resolution (256x160) (\textit{appearance not preserved, edge pixelization})}}}
  \label{fig:p2f_ablation}
\end{figure*}

\section{Experiments}
\textbf{Datasets.}
Our networks are trained on cropped images, each containing a single person. 
The VIP dataset~\cite{zhou2018adaptive} is used to train the P2B network. The dataset contains 404 densely annotated videos with pixel-wise semantic part categories and a total of $21$k frames. After cropping each separate person, the customized dataset contains a total of $62$k images.
The B2F network is trained by combining two datasets. MHPv2 ~\cite{li2017multiple}
contains $25$k images with an average of three people per image. After removing small and highly occluded people, $53$k unique people remain. 
CIHP \cite{gong2018instance} contains $28$k images. After pre-process, $1.7$k different people with a total of $44$k images (average of 25 images per person) remain. For each person, up to 15 random pairs are chosen, resulting in $19$k unique pairs. Additional implementation details are provided in the supplementary.

For the numerical analysis, {{the target is taken from the driving video, establishing a valid ground-truth. For visual comparisons, where no ground-truth is required,}} we select 21 target images, out of which 11 are clearly visible, in a full-bodied frontal pose (denoted as the ``simple'' targets). 
Ten target images depict individuals who are not fully visible, or not in a standing frontal pose, denoted as the ``challenging'' targets. {{All target images used are provided in the supplementary.}} 
The vast majority of the selected target images are taken out of the DFDC dataset~\cite{dolhansky2020deepfake}. The DFDC dataset is uniquely diverse, allowing a comprehensive evaluation of the methods over different attributes, such as ethnicity, gender and age, but also pose, viewpoint and scale. Additional images were obtained from consenting individuals, {{attached as part of the supplementary}}.

\smallskip\noindent{\bf Baselines.}
We compare our results with state of the art methods that represent the different approaches existing in the literature for the task of dance generation. 

When available, we use the authors' pre-trained weights; otherwise, we train the models with our dataset, following the authors' instructions.
\textbf{fsV2V}\cite{wang2019fewshotvid2vid} generates the entire video using a target image, OpenPose and DensePose data. It employs a hyper-network that predicts the weights of a vid2vid network. To achieve improved results, we followed the authors' instructions and fine-tuned the network for each video.
\textbf{Pose Warping}\cite{balakrishnan2018synthesizing} generates a new frame by transforming each body part of the target, based on pose keypoints of the source and target images, followed by a fusion operation.
\textbf{SPT}\cite{song2019unsupervised} resembles our approach, as the generator consists of two main parts. The first, a semantic generator, generates a new semantic map based on the source semantic segmentation and the new pose. The second, an appearance generator, renders the final frame. Generation is performed gradually in 128x128 and 256x256 pixels. Since the authors did not release the code for their semantic generator, we employ our P2B results instead.

\smallskip\noindent{\bf Evaluation metrics.} 
All comparisons are made over targets and driving videos that do not appear in any training datasets. We use nine videos with an average of 300 frames each, obtained with consent from a video blogger. The evaluation metrics can be naturally divided into two distinct groups: quality and pose similarity. 

For pose similarity, DPBS (DensePose Binary Similarity) and DPIS (DensePose Index Similarity) calculations~\cite{gafni2020wish} are used and are further adapted to serve as  semantic segmentation similarity metrics (SSBS and SSIS). DPBS (SSBS) evaluates the IoU between a binary representation of the ground-truth and generated DensePose (the HP network), while DPIS (SSIS) evaluates the mean over each body-part index, for the same network. 

For quality metrics, we rely on SSIM~\cite{wang2004image}, LPIPS~\cite{zhang2018perceptual} and FID~\cite{heusel2017gans} to capture perceptual notions. LPIPS is applied with both the VGG~\cite{simonyan2014very} and SqueezeNet~\cite{iandola2016squeezenet} networks. 

In addition, a user study is conducted among $n=50$ participants. Each participant is shown the nine videos, where each video is shown as an instance generated by our method alongside an instance generated by one of the previous methods. The videos and targets are randomly selected such that three videos are presented for each method. The participant is asked to then select the video they prefer for each of the nine pairs of videos shown.

\subsection{Results} Since the baseline methods struggle with challenging conditions, we measure  performance only on the ``simple'' settings. As can be seen in Tab.~\ref{tab:results}, our method achieves superior results over all baselines and metrics. Those are apparent for both pose similarity and quality metrics. Additionally, the users present an overwhelming preference towards our method. 

A visual comparison can be seen in Fig.~\ref{fig:previous1} and in the supplementary (image and video samples). For both ``simple'' and ``challenging'' targets, our results are noticeably better at appearance preservation and quality.

\subsection{Ablation}
{A visual ablation study is provided, where a distinction is made between structural and full pipeline aspects. The necessity of certain components in B2F and the existence of the FR network are examined with details in Fig.~\ref{fig:p2f_ablation}, while P2B is evaluated in in the supplementary. 
For each case, the dominant discrepancies are emphasized in a green square for our result, and a red square for each ablation case. }

\section{Conclusions}

The desiderata of person animation techniques include not just visual quality, natural motion, motion  fidelity, and appearance preservation, but also the ability to capture multiple body types, gender, ethnicity, and age groups. 
Diversity in human pose generation is imperative to making sure technology is inclusive and can benefit everyone. However, it is often neglected in the literature.

The method we present, provides a much more detailed model of the human body, its appearance and its motion, than previous approaches. It is trained in a way that encourages it to address diverse inputs. In a comprehensive set of experiments, we demonstrate that the method is able to obtain better visual quality and better fidelity of both motion and appearance than the existing methods.

\section*{Acknowledgments}
The authors would like to thank Bao Tran from 'Learn How To Dance' for allowing us to use his videos for inference.

{\small
\bibliographystyle{ieee_fullname}
\bibliography{dance}
}
\clearpage
\appendix

\section{Additional results}
\begin{figure}
  \centering
    \begin{tabular}{c@{}c@{}c} 
 \includegraphics[width=3cm]{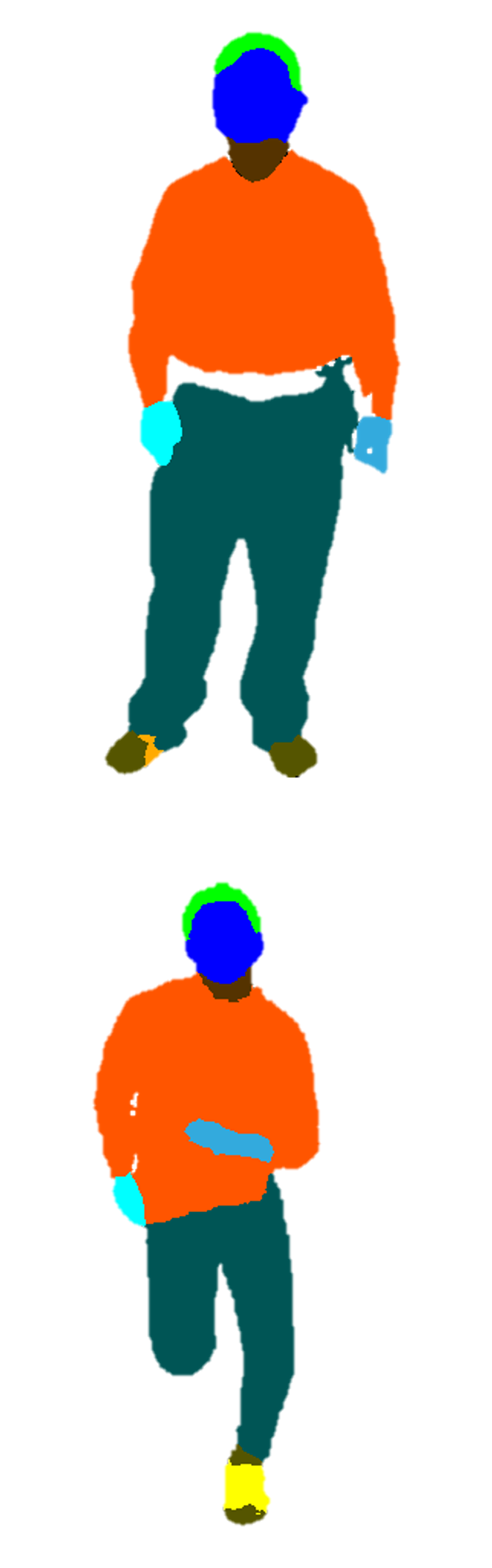} & 
 \includegraphics[width=3cm]{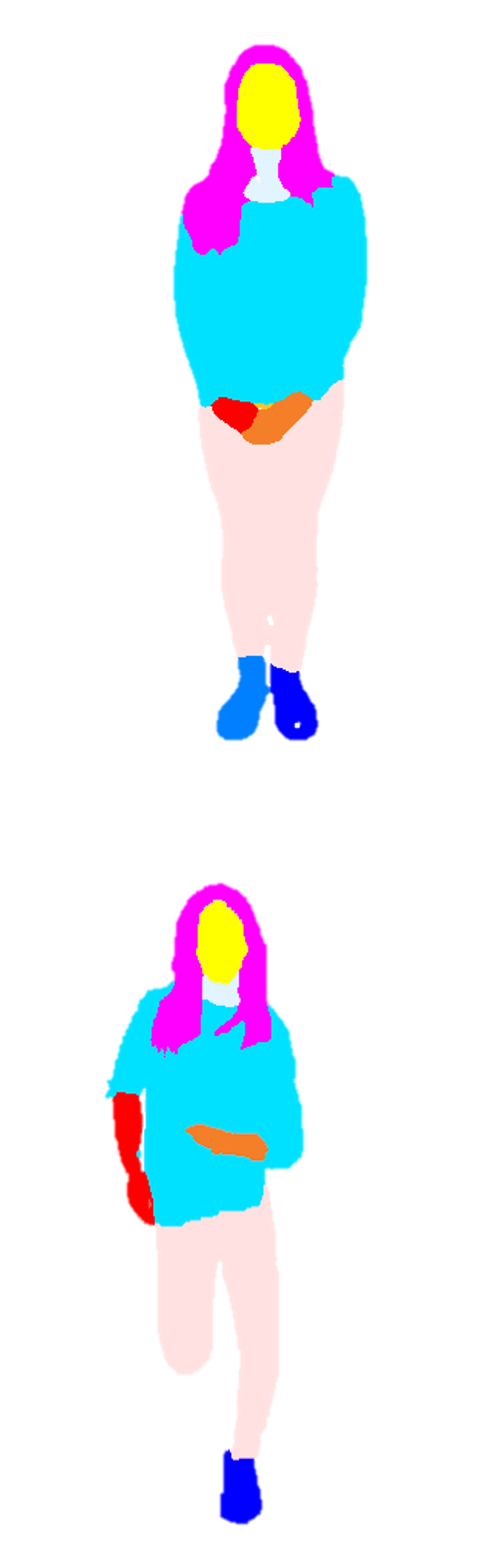} & 
 \includegraphics[width=3cm]{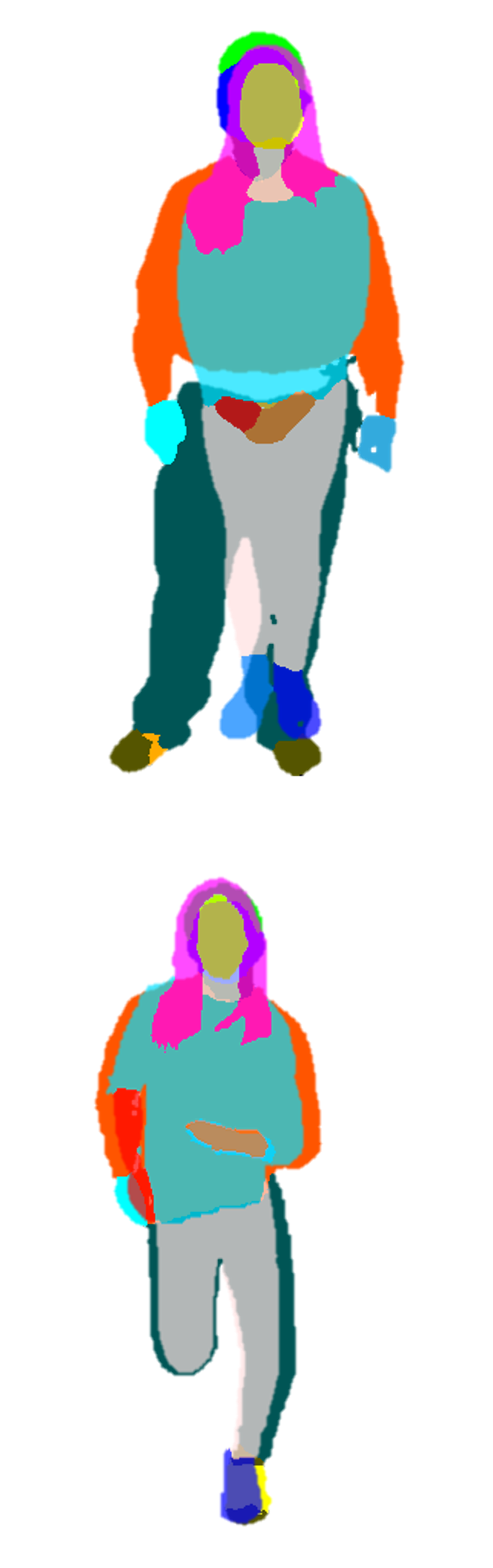} \\
 (a) & (b) & (c) \\
 \end{tabular}
\caption{Body structure diversity example. For the same driving pose, two generated individuals are evaluated. The body structure, as captured by the semantic segmentation of the target images (row 1) for the first (a) and second (b) person, can be see to be distinct, as emphasized by overlaying one over the other (c). The distinction in body structure can be seen to be maintained in the corresponding rendered images (row 2).}
\label{fig:body_type}
\end{figure}

\textbf{Body diversity.}
As mentioned, explicit augmentations encourage diverse body structure preservation. Fig.~\ref{fig:body_type} showcases this aspect, where two individuals are chosen with distinctively different body structures. The semantic maps of both individuals are shown in the first row, while the generated semantic map for the same pose is shown in the second row. The individuals are overlaid in column (c) for clarity.

\begin{figure*}
  \centering
 \includegraphics[width=1.0\textwidth]{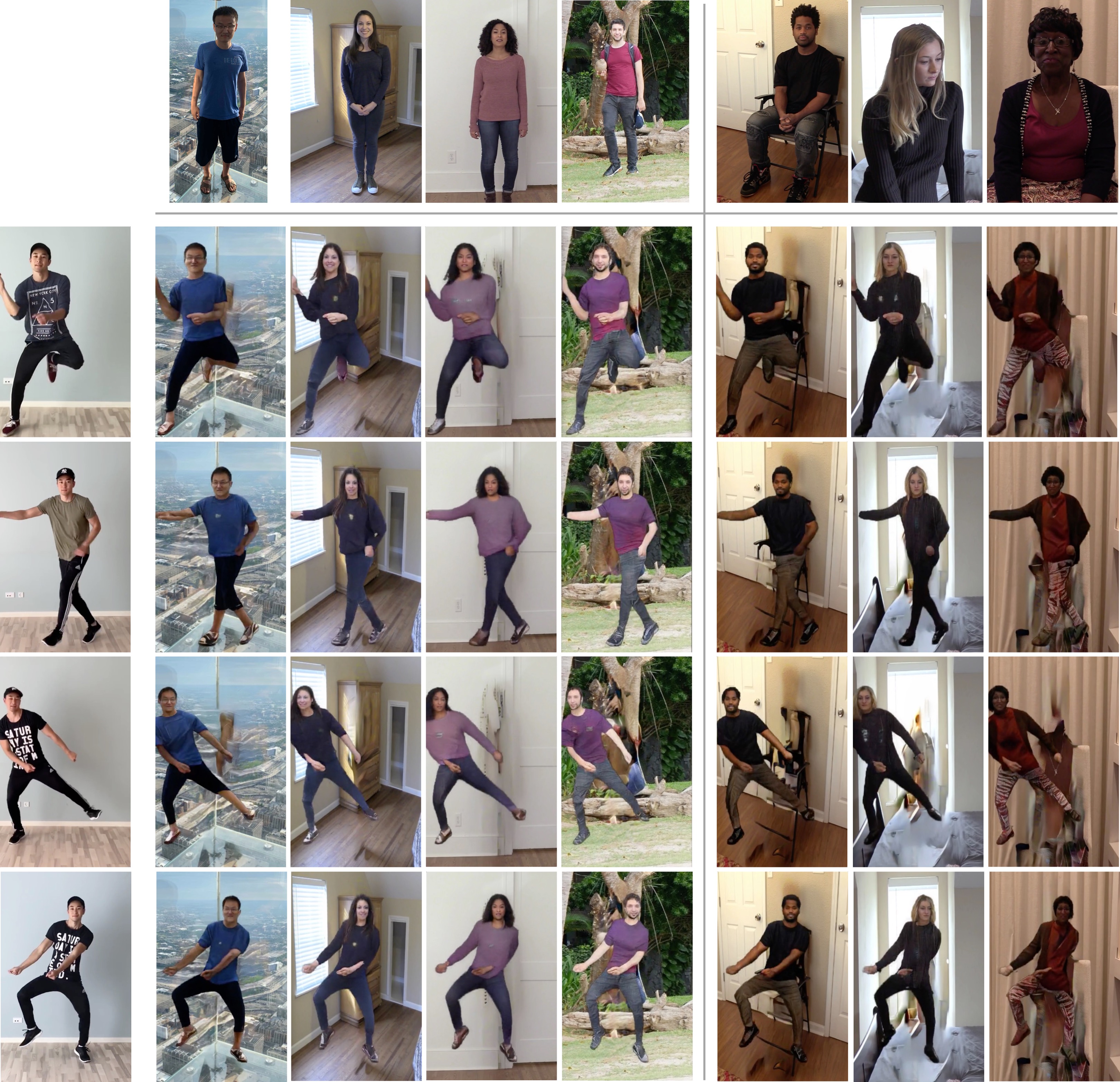}
  \caption{Sample results. Four "simple" and three "challenging" targets are shown. In all cases, realistic samples are rendered for a diverse set of appearances and poses. Additional results can be seen in the accompanying video. Note that the facial expression is transferred from the target image, rather than from the driving image.
  }
  \label{fig:sample1}
\end{figure*}

\textbf{Sample results.} Additional results are provided in Fig.~\ref{fig:sample1} for both "simple" and "challenging" target images, over different poses. In all cases, realistic samples are rendered.

\begin{figure*}
\medskip
  \centering
\begin{tabular}{@{}c@{~}c@{~}c@{~}c@{~}c@{~}c@{}}
  \includegraphics[height=3.5cm]{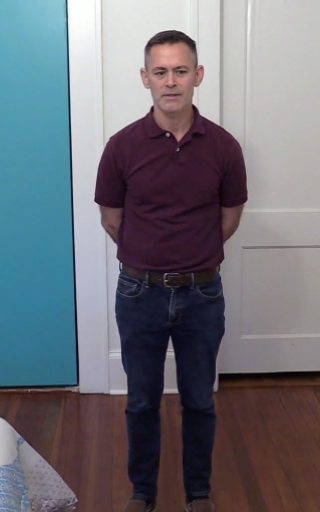} & \includegraphics[height=3.5cm]{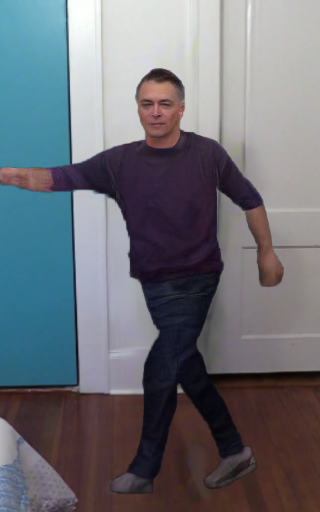} & \includegraphics[height=3.5cm]{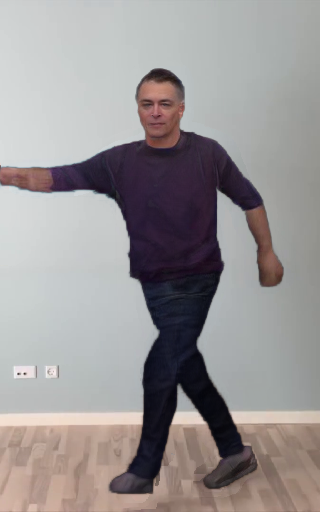} & \includegraphics[height=3.5cm]{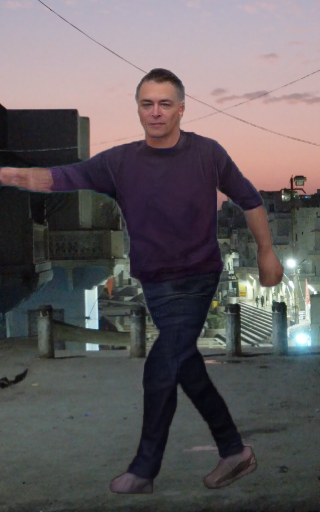} & \includegraphics[height=3.5cm]{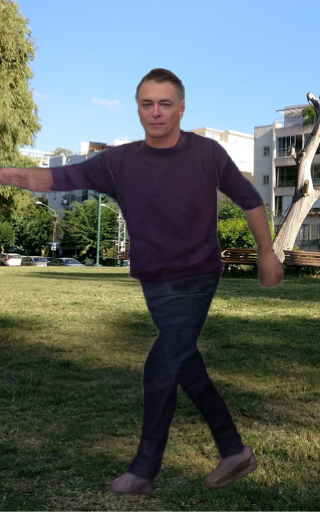} &
  \includegraphics[height=3.5cm]{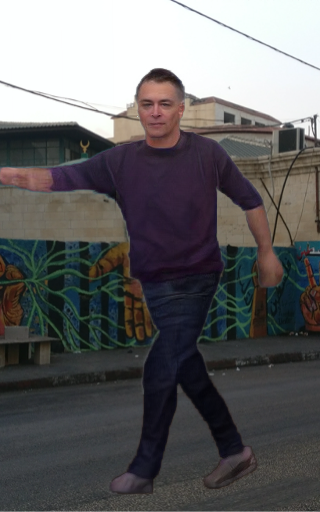} \\
    \includegraphics[height=3.5cm]{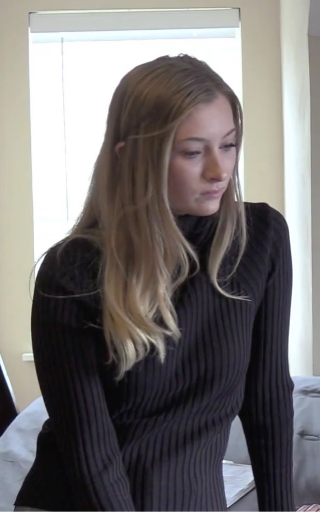} & \includegraphics[height=3.5cm]{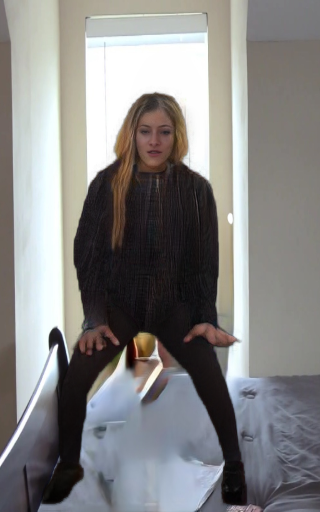} & \includegraphics[height=3.5cm]{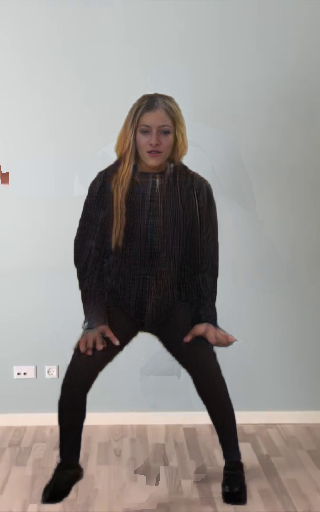} & \includegraphics[height=3.5cm]{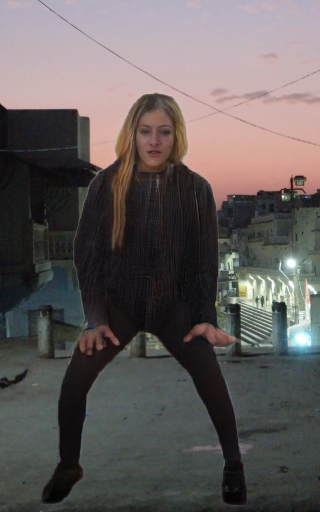} & \includegraphics[height=3.5cm]{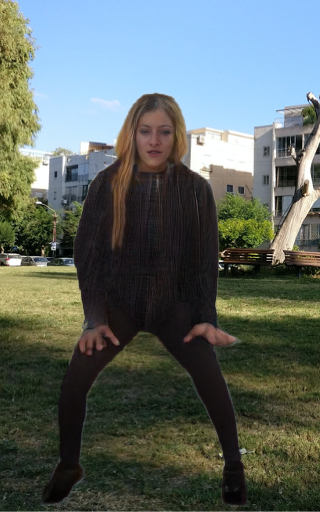} &
    \includegraphics[height=3.5cm]{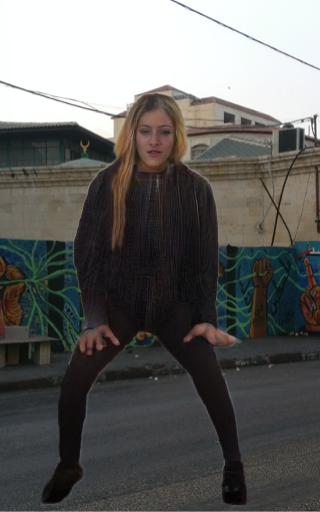}
 \\
  (a) & (b) & (c) & (d) & (e) & (f)
  \end{tabular} 
  %
  \caption{Interchangeable backgrounds. The generated blending mask is used to seamlessly embed the rendered person into any given background. {{(a) Target image, (b) embedded into the inpainted target background (c) embedded into the inpainted driving video background (residual shadows complement the naturalness of the embedded person), (d)-(f) embedded into various backgrounds.}}}
  \label{fig:backgrounds}
\end{figure*}

\textbf{Interchangeable backgrounds.} Generating a blending mask is an integral part of the method, as it enables embedding the generated person into any background. Fig.~\ref{fig:backgrounds} demonstrates this ability. As seen in column (c), by embedding the rendered person back into the inpainted source video, the shadows of the original dancer complement the naturalness of the rendered person.

\section{Additional Comparison}
\begin{figure}[t]
\begin{center}
   \includegraphics[width=0.95\linewidth]{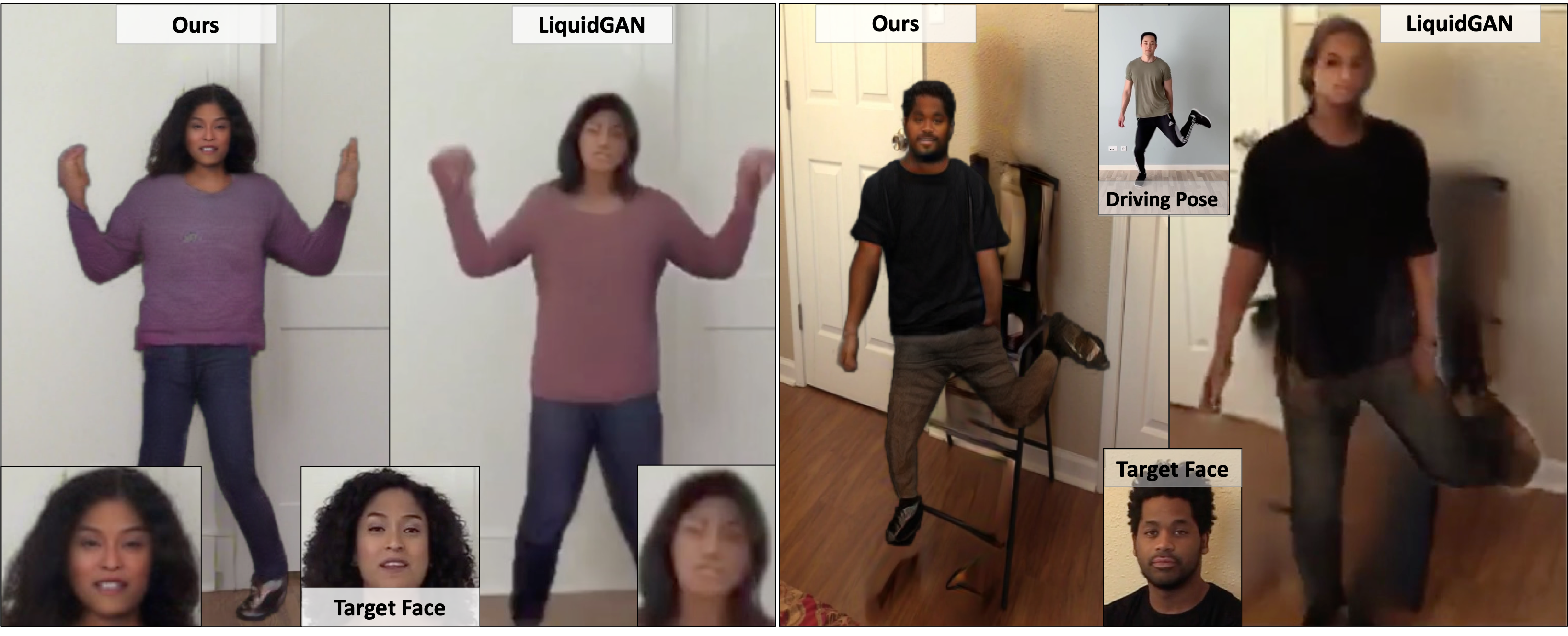}
\end{center}
   \vspace{-3mm}
   \caption{Ours vs. LiquidGAN. (L) Easy, (R) challenging targets.}
   \vspace{-5mm}
\label{fig:liquidgan}
\end{figure}

{{Comparison with Liquid-GAN~\cite{lwb2019} is presented in Fig.~\ref{fig:liquidgan}. Compared to ~\cite{lwb2019}, our biggest advantage is natural motion, which cannot be conveyed here. As shown in Fig.~\ref{fig:liquidgan}, our method also surpasses in terms of resolution, appearance, pose, and background replacement.}}

\section{Additional implementation details}
The P2B and B2F networks are trained with the \textit{ADAM}\cite{adam} optimizer applying a learning rate of $0.0002$ and ${(\beta 1,\beta 2)=(0.5,0.999)}$. The P2B is trained for 280 epochs, with a batch size of 128, while the B2F is trained for 60 epochs, with a batch size of 32. The Face Refinement network is trained with the same optimizer, a learning rate of $0.0001$, ${(\beta 1,\beta 2)=(0.5,0.999)}$, for 40 epochs and a batch size of 256.

\section{Limitations}
Our method is driven by pose representations, and conditioned over a semantic map of the target person. As previous methods, ours as well suffers from a strong dependency on the quality of the detected driving pose, though is somewhat robust to the conditioned semantic map (hence capable of handling "challenging" targets).

Body structure preservation is an important aspect of dance reenactment, and receives significant attention in this work. Although this method is able to preserve some body structure, it is still constrained by the strong bias that accompanies datasets used to train the different networks, specifically the Pose2Body network.

The rendered blending mask enables to seamlessly blend the generated person into any given background, yet does not provide a complete solution for all environmental surroundings, such as shadows. A partial resolution for this gap is using the inpainted source video as the background, as seen in Fig.~\ref{fig:backgrounds}(c) and in the accompanying video.

\begin{figure*}
  \centering
 \includegraphics[width=0.85\textwidth]{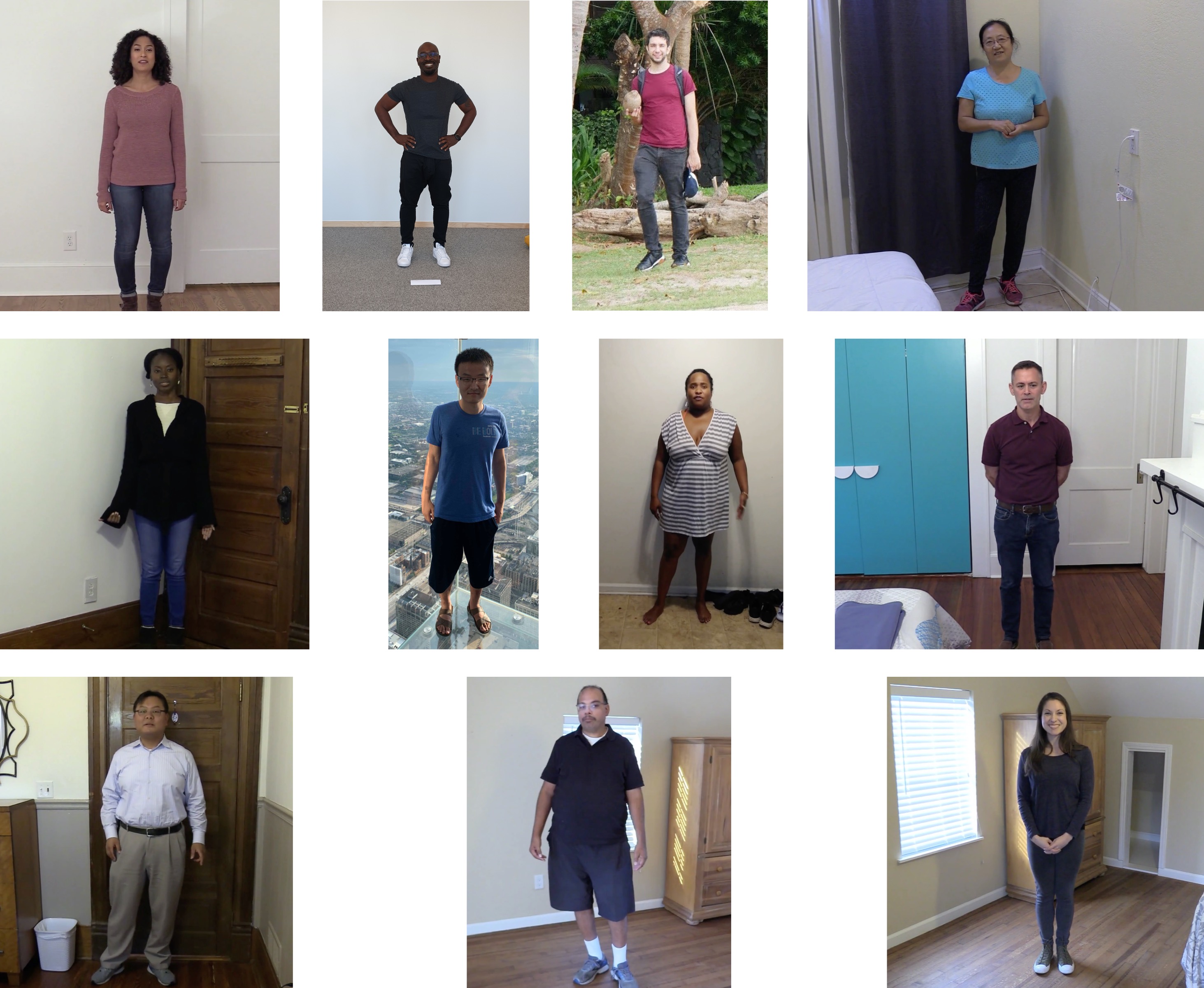}
  \caption{"Simple" targets used for human preference survey and visual comparison. }
  \label{fig:easy_targets}
\end{figure*}

\begin{figure*}
  \centering
 \includegraphics[width=0.8\textwidth]{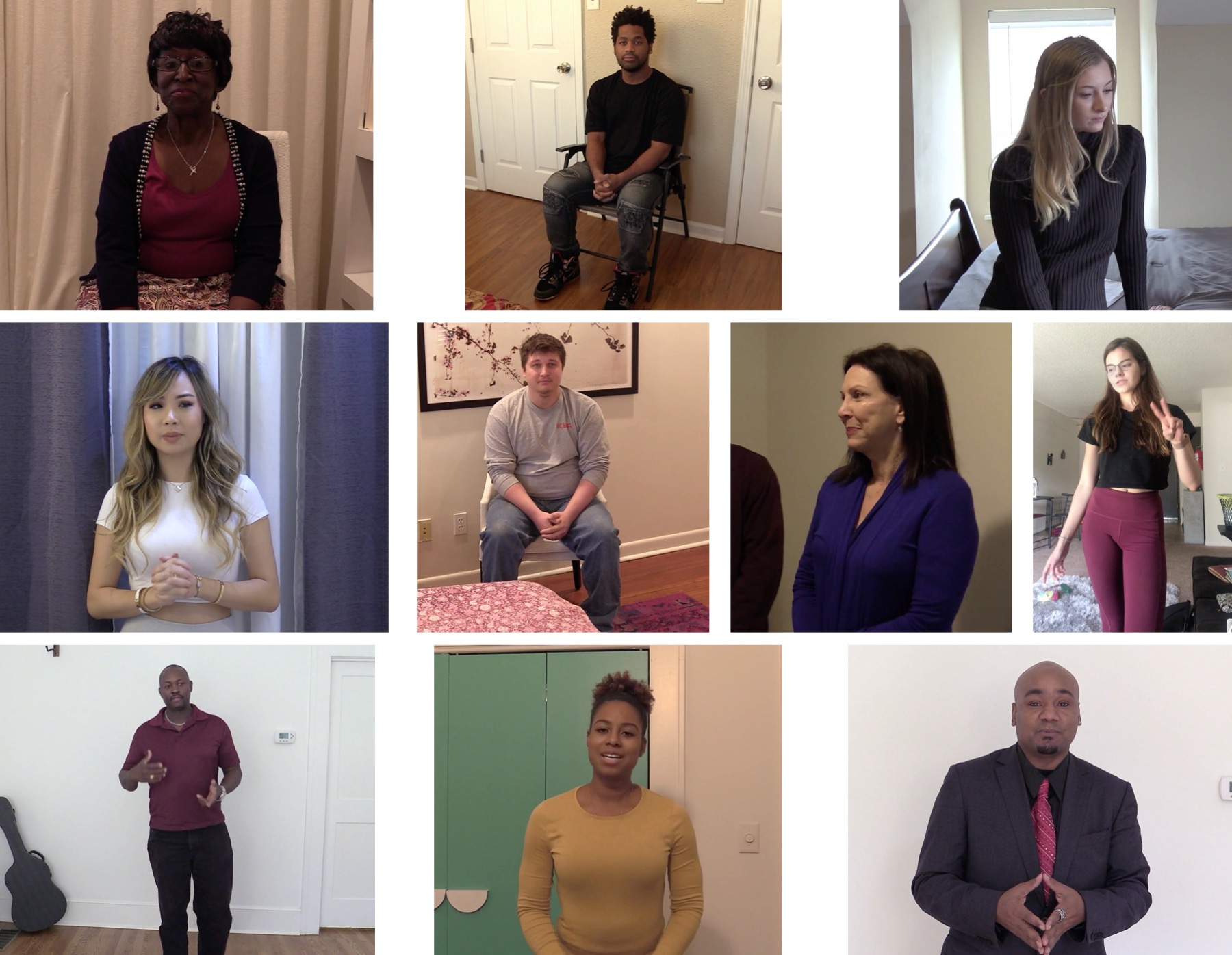}
  \caption{"Challenging" targets used for visual comparison.}
  \label{fig:challenging_targets}
\end{figure*}

\section{P2B ablation experiment.}

\begin{figure*}[t]
\medskip
  \centering
\begin{tabular}{@{}c@{~}c@{~}c@{~}c@{~}c@{}}
  \includegraphics[height=3.5cm]{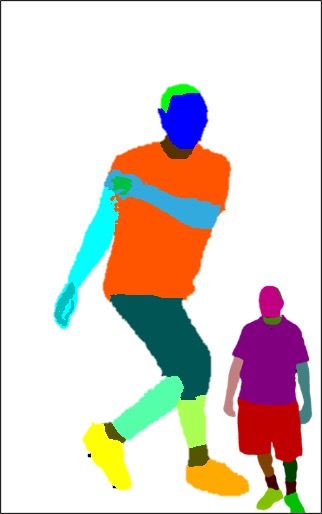} & \includegraphics[height=3.5cm]{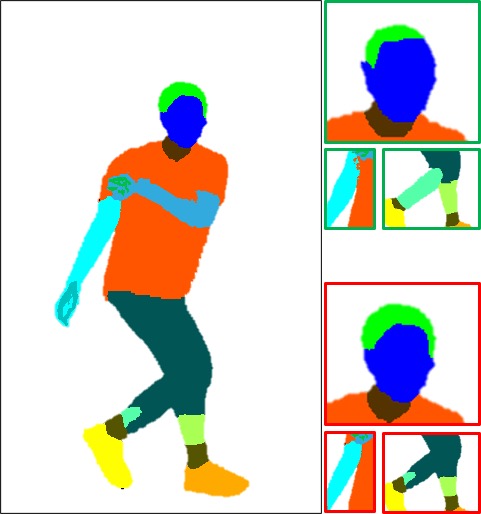} & \includegraphics[height=3.5cm]{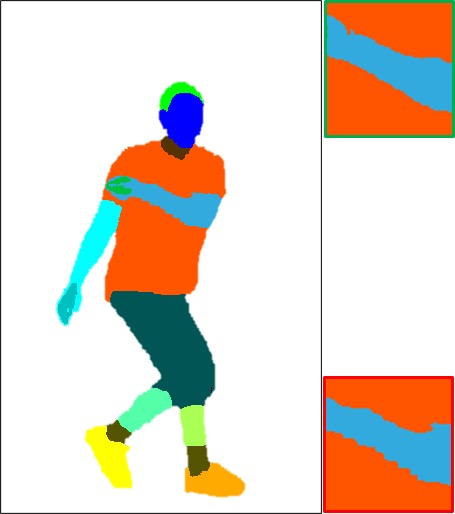} & \includegraphics[height=3.5cm]{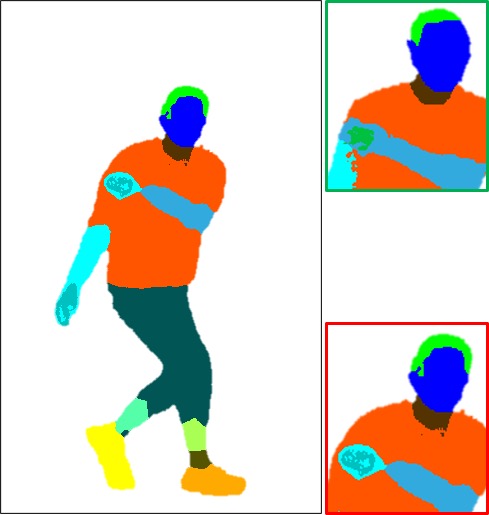} & \includegraphics[height=3.5cm]{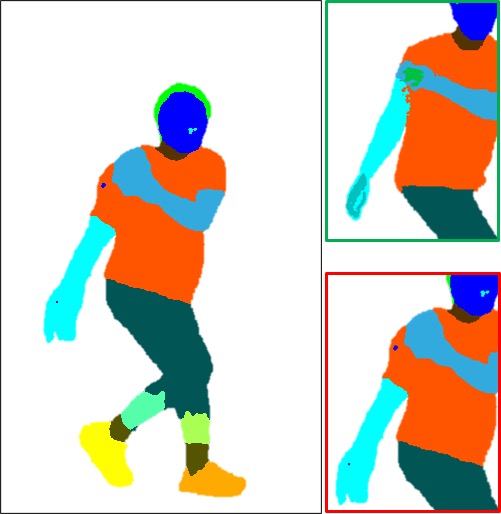} 
 \\
  (a) & (b) & (c) & (d) & (e)
  \end{tabular} 
  %
  \caption{P2B ablations. {{(a) Our result and the target parsing (scaled down). The following are various variants. In red, a zoom in version, and in green the same zoom applied to the output of the full method. (b) No squeezing and stretching of the input/output parsing (\textit{body structure, hair, and clothing less consistent}), (c) a less accurate version of DensePose is used (\textit{boundary artifacts}), (d) DensePose is not used as input (\textit{increased limbs artifacts, instability in body structure}), (e) no DP and no hand/finger labels (\textit{enormous arms}).}}}
  \label{fig:p2p_ablation}
\end{figure*}

The ablation experiment for the P2B network is presented in Fig.~\ref{fig:p2p_ablation}. We highlight dominant discrepancies by a green square for our result and a red square for each ablation case.

\section{Quantitative ablation}
{{We focus on a qualitative ablation for the following reasons: (1) As the main objective is rendering a novel person, real dance generation does not have a ground-truth, making majority of the metrics irrelevant (e.g. disentangling the body structure from the driving pose is not relevant, resulting in deceptively better results for the ablation case), (2) numerical metrics often hide the real impact of losses trade-offs. As an example, we achieve better LPIPS if we do not use any face-related losses, as the addition of a face-related loss adds conflicting considerations. However, face appearance is very important in human perception. 
Nevertheless, quantitative results are presented in Tab.~\ref{tab:quant_abl}. As expected, it shows a trade-off between the losses, e.g., removing the face-related losses hurts face perception significantly, while slightly improving other metrics.}}

\begin{table*}
\begin{center}
\begin{tabular}{|l|c|c|c|c|c|c|c|c|}
\hline
Method & SSBS $\uparrow$ & SSIS $\uparrow$& DPBS $\uparrow$& DPIS $\uparrow$& LPIPS 	$\downarrow$& LPIPS $\downarrow$&  SSIM $\downarrow$ & FID $\downarrow$ \\
       &      &      &      &      & (VGG) &(SqzNet)&      &  \\
\hline\hline
(P2B) No Squeeze/Stretch  & $0.898$ & $\textbf{0.220}$ & $0.926$ & $\textbf{0.514}$ & - & - & - & - \\
(P2B) No Accurate DP  & $\textbf{0.902}$ & $0.218$ & $0.927$ & $0.500$ & - & - & - & - \\
(P2B) No DP  & $0.869$ & $0.197$ & $0.884$ & $0.460$ & - & - & - & - \\
(P2B) No Fingers/DP  & $0.869$ & $0.197$ & $0.884$ & $0.460$ & - & - & - & - \\
(B2F) No FR & $0.873$ & $0.208$ & $0.896$ & $0.468$ & $0.378$ & $0.299$ & $0.133$ & $\textbf{70.880}$ \\
(B2F) No Mask & $0.873$ & $0.216$ & $0.891$ & $0.458$ & $0.379$ & $0.300$ & $0.135$ & $74.503$ \\
(B2F) No Fingers  &  $0.863$ & $0.208$ & $0.897$ & $0.467$ & $0.375$ & $0.296$ & $0.130$ & $73.715$ \\
(B2F) No Face-loss/LR  & $0.873$ & $0.217$ & $0.896$ & $0.465$ & $\textbf{0.373}$ & $0.293$ & $0.128$ & $77.032$ \\
Ours & $\textbf{0.902}$ & $0.218$ & $\textbf{0.928}$ & $0.500$ & $0.375$ & $\textbf{0.283}$ & $\textbf{0.116}$ & $83.95$  \\

\hline
\end{tabular}
\end{center}
\caption{{{Quantitative ablation. }}}
\label{tab:quant_abl}
\end{table*}

\section{Inference time}
{{Inference time considerations mainly focus on mitigating bottlenecks and maximum parallelization. The main bottlenecks are currently the DensePose and B2F networks' run-time. To achieve real-time inference, we would either remove DP, or employ DP on a low-resolution image. Reducing the B2F run-time could be achieved by a range of optimizations, such as reducing channel number, or converting ResSPADE blocks to lighter ResBlocks (e.g. MobileNetV3). This results with the sequence of (1) OP+DP, (2) P2B, (3) B2F, (4) FR (the rest is done once per person, and could be pre-processed). As we do not employ any temporal components, each of the 5 networks could run in parallel on 5 GPUs (after passing the first 4 frames). This would bring us to approx. (1) 41ms, (2) 20ms, (3) 20ms, (4) 30ms, where (1) is the limiting factor, resulting in 24FPS (can be improved by adding an additional GPU for OP), with a latency of 111ms.}}

\section{Region refinement}
{{The face refinement utilized a network trained specifically on faces to improve quality and appearance. 
In a similar manner to face refinement, it is possible to add losses emphasizing each part of interest (e.g. hands, shirt, pants), utilizing a specific network (e.g. trained on hands) or a general one (e.g. ImageNet). This is already done implicitly through the pre-trained encoder, yet explicit losses (as done for the face part) can provide additional improvement.}}

\end{document}